\documentclass[sigconf, nonacm]{acmart}

\newcommand\vldbdoi{XX.XX/XXX.XX}
\newcommand\vldbpages{XXX-XXX}
\newcommand\vldbvolume{16}
\newcommand\vldbissue{4}
\newcommand\vldbyear{2022}
\newcommand\vldbauthors{\authors}
\newcommand\vldbtitle{\shorttitle} 
\newcommand\vldbavailabilityurl{https://github.com/gunduzvd/Scalable-Graph-Convolutional-Network-Training-on-Distributed-Memory-Systems}
\newcommand\vldbpagestyle{empty} 

\usepackage[utf8]{inputenc} 
\usepackage[T1]{fontenc}    
\usepackage{hyperref}       
\usepackage{url}            
\usepackage{booktabs}       
\usepackage{amsfonts}       
\usepackage{nicefrac}       
\usepackage{microtype}      
\usepackage{xcolor}         
\usepackage{graphicx}
\usepackage{colortbl}

\usepackage[a-1b]{pdfx}

\usepackage{amsmath}
\usepackage{fixmath}
\usepackage{multirow}
\usepackage{pgfplots}
\usepackage{tikz}
\usepgfplotslibrary{groupplots}
\pgfplotsset{compat=1.15}

\usepackage[ruled,vlined,noend,linesnumbered]{algorithm2e}

\usepackage{subcaption}
\usepackage{titlesec}
\titleformat*{\subsubsection}{\large\bfseries}

\newcommand{\mat}[1]{\mathbold{#1}}
\newcommand{\mA}{\mat{A}}
\newcommand{\mH}{\mat{H}}
\newcommand{\mW}{\mat{W}}
\newcommand{\mD}{\mat{D}}
\newcommand{\mZ}{\mat{Z}}
\newcommand{\mX}{\mat{X}}
\newcommand{\mG}{\mat{G}}
\newcommand{\mS}{\mat{S}}

\newcommand{\Gg}{\mathcal{G}}
\newcommand{\Hg}{\mathcal{H}}
\newcommand{\Ns}{\mathcal{N}}
\newcommand{\Es}{\mathcal{E}}
\newcommand{\Vs}{\mathcal{V}}
\newcommand{\Ss}{\mathcal{S}}
\newcommand{\Rs}{\mathcal{R}}

\begin{document}


\title{Scalable Graph Convolutional Network Training on Distributed-Memory Systems}


\author{Gunduz Vehbi Demirci}
\authornote{\textit{Previously at the University of Warwick. This publication describes work performed at the University of Warwick and is not associated with Imagination Technologies.}}
\affiliation{%
  \institution{Imagination Technologies}
  \country{United Kingdom}
}
\email{gunduz.demirci@imgtec.com}

\author{Aparajita Haldar}
\affiliation{%
  \institution{University of Warwick}
  \country{United Kingdom}
}
\email{aparajita.haldar@warwick.ac.uk}

\author{Hakan Ferhatosmanoglu}
\authornote{\textit{Also with Amazon Web Services. This publication describes work performed at the University of Warwick and is not associated with Amazon.}}
\affiliation{%
  \institution{University of Warwick}
  \country{United Kingdom}
}
\email{hakan.f@warwick.ac.uk}

\begin{abstract}
Graph Convolutional Networks~(GCNs) are extensively utilized for deep learning on graphs. 
The large data sizes of graphs and their vertex features make scalable training algorithms and distributed memory systems necessary.
Since the convolution operation on graphs induces irregular memory access patterns, designing a memory- and communication-efficient parallel algorithm for GCN training poses unique challenges.
We propose a highly parallel training algorithm that scales to large processor counts.
In our solution, the large adjacency and vertex-feature matrices are partitioned among processors. 
We exploit the vertex-partitioning of the graph to use non-blocking point-to-point communication operations between processors for better scalability.
To further minimize the parallelization overheads, we introduce a sparse matrix partitioning scheme based on a hypergraph partitioning model for full-batch training. We also propose a novel stochastic hypergraph model to encode the expected communication volume in mini-batch training.
We show the merits of the hypergraph model, previously unexplored for GCN training, over the standard graph partitioning model which does not accurately encode the communication costs.
Experiments performed on real-world graph datasets demonstrate that the proposed algorithms achieve considerable speedups over alternative solutions. The optimizations achieved on communication costs become even more pronounced at high scalability with many processors. The performance benefits are preserved in deeper GCNs having more layers as well as on billion-scale graphs.





\end{abstract}

\maketitle

\section{Introduction}
\label{sec:intro}

Graph Convolutional Networks~(GCNs) generalize the convolution operation, performed by convolutional neural networks on structured data~(e.g., images, time-series), to graphs~\cite{micheli2009neural,shuman2013emerging}.
GCNs are used in a wide range of data intensive graph applications such as node classification~\cite{kipf2016semi,miao2021lasagne}, traffic forecasting on road networks~\cite{yu2017spatio}, and recommender systems on user-item graphs~\cite{ying2018graph}.


While graph-based learning models have been highly successful, the scale of large graphs, including their multi-dimensional features for the vertices, necessitates the use of distributed memory systems~\cite{tripathy2020reducing,zheng2020distdgl,ma2019neugraph,jia2020improving}.
During the feedforward and backpropagation phases in GCN training, the graph convolution operation involves message passing and aggregation steps that induce irregular data accesses due to complex graph inter-connectivity.
Existing systems use graph partitioning algorithms designed for traditional graph algorithm workloads (e.g., connected components, shortest paths), which do not take complex GCN data access patterns into consideration.
Therefore, intelligent message passing strategies need to be employed to achieve a communication-efficient distributed-memory parallel inference and training solution.


Sparse-dense matrix multiplication~(SpMM) and dense matrix multiplication~(DMM) are core kernel operations in GCN training.
SpMM achieves convolution whereas DMM corresponds to propagating vertex-feature vectors through a single layer neural network.
There have been improved solutions proposed for parallel SpMM~\cite{selvitopi2021distributed,koanantakool2016communication,kunchum2017improving} and DMM~\cite{ballard2012communication} problems. However, the special requirements of combining SpMM and DMM for scalable GCN training and ensuring efficient forward propagation and backpropagation phases in GCNs remain under-explored. In particular, communications incur high latency and bandwidth costs to aggregate feature matrices during feedforward as well as to aggregate gradients and update parameter matrices during backpropagation.

While recent parallel/distributed algorithms achieve GCN training for GPU clusters and cloud systems~\cite{zheng2020distdgl,tripathy2020reducing}, these typically perform broadcast- and allreduce-type of collective communication operations.
Sparse data communication and compression methods have been considered to alleviate the scalability issues of allreduce for larger models and processor counts~\cite{demirci2021partitioning, fei2021efficient,kostopoulou2021deepreduce,lin2017deep}.
However, such redundant data and message transfer causes unnecessary communication overheads.
Moreover, in GCN training, model parameter matrices are significantly smaller than the adjacency and vertex-feature matrices, so performance improvements in allreduce communication are not significant in the overall parallel execution time.
Instead, efficient parallelization of SpMM performed on the large graph data can lead to higher performance increase.
For example, Sancus~\cite{peng2022sancus} is a recent model that adaptively avoids broadcast communications to reduce network traffic in data-parallel GNNs.
However, parallel SpMM still requires broadcast-type communication, which is the main performance bottleneck in GCN training, due to its high memory and bandwidth costs.
A parallel algorithm, CAGNET~\cite{tripathy2020reducing}, performs broadcasts among processors turn-wise to transfer vertex-features in small portions but suffers significant latency overheads.
Therefore, a viable alternative is to design a mechanism that can utilize non-blocking point-to-point communications that move only the necessary data among processors.




We introduce a highly parallel algorithm for training GCNs on distributed-memory systems. 
Our solution achieves scalability by replacing the blocking broadcast communications in existing approaches with non-blocking point-to-point communications for parallel SpMM and transferring only the necessary data with minimal number of messages between processors. 
The solution employs a one-dimensional (1D) partitioning on large adjacency, vertex-feature, and gradient matrices for parallel SpMM computations in feedforward and backpropagation phases. It replicates parameter matrices across processors due to their relatively smaller sizes.
This enables data locality for performing DMM computations without any communication. Allreduce communication is needed for aggregating gradients, which has a negligible cost compared to the communication costs incurred in parallel SpMM. 


The use of point-to-point communication operations in our solution enables communication to be reduced further via sparse matrix partitioning strategies~\cite{chen2006mpipp}.
We develop a sparse matrix partitioning scheme to distribute the adjacency, vertex-feature, and gradient matrices used in computations among processors, based on a hypergraph partitioning model for the original graph. We show that the hypergraph partitioning model encodes SpMM communication costs more accurately than the graph partitioning model which is in popular use (e.g., in DistDGL~\cite{zheng2020distdgl}).
To capture the randomness when communication operations are performed for mini-batch training instead of full-batch training, we also introduce a novel stochastic hypergraph model. This model encodes the \emph{expected} communication volume in parallel mini-batch training and can be utilized for any mini-batch sampling strategy.


We focus on large-scale CPU clusters, commonly used for big sparse problem instances in scientific computing, since research towards adapting existing, relatively inexpensive supercomputing systems towards deep learning is gaining attention. 
We also demonstrate our proposed solution 
on GPU clusters 
by replacing local computations with GPU kernels and using NCCL~\cite{awan2016efficient}. 

We perform extensive experiments on real-world network datasets.
Experimental results show that our solution is highly efficient and scales to large processor counts.
We show that the proposed distributed solution achieves considerable speedups over the single-node GCN implementation in Deep Graph Library~(DGL)~\cite{wang2019deep}, especially on large graphs having low average node degrees. 
Our hypergraph model generally outperforms the graph model as it correctly encodes the tasks and data dependencies by exploiting sparsity in connectivity patterns. 
The time spent on communication operations decreases with the increasing number of processors. 
This provides a scalability advantage over current alternatives that use inefficient collective communications. 
Using the novel stochastic hypergraph partitioning algorithm, we achieve further reductions in the communication volume for mini-batch training.

A summary list of contributions of this paper is given below:
\begin{itemize}
    \item We propose a highly parallel GCN training algorithm that exploits sparsity in communications and data locality in computations to scale the training process.
    \item We show the merits of our hypergraph-based data partitioning scheme over the more popular graph-based approach for distributed full-batch training of GCNs, to reduce communication overheads while satisfying load balance. 
    \item We propose a novel stochastic hypergraph partitioning model which can be utilized 
    in parallel mini-batch training.
    \item On a set of real-world graph datasets (e.g., citation graphs, social networks, road networks, co-purchasing networks), we evaluate the performance of the proposed algorithm and data partitioning models, and provide further insights for scalable data processing and training for GCNs.
\end{itemize}

\section{Related Work}
\label{sec:rel}


\subsection{Distributed Graph Processing}

Distributed systems have been widely employed for graph analytics, from parallel processing to streaming graph updates and cloud-based graph engines~\cite{malewicz2010pregel, gonzalez2012powergraph, ko2018turbograph++,shao2013trinity,karimi2019gpu,sun2017graphgrind}.
Several graph analytics APIs, such as GraphX~\cite{xin2013graphx}, are built atop Apache Spark or similar frameworks. These systems face CPU utilization bottlenecks that can be avoided with our data parallelization solution, enabling better scaling. 
In contrast to vertex-centric models, which suffer high communication overheads~\cite{lu2014large,han2014experimental}, graph- and block-centric models utilize local graph partition structure to reduce communication and scheduling~\cite{tian2013think,yan2014blogel}.
Our parallel solution instead exploits sparse connectivity patterns to achieve better data locality, enabling efficient communication across thousands of compute nodes. 




Graph partitioning is widely employed for improving the efficiency of different types of queries~\cite{zhao2013partition,sarwat2013horton+,fan2022application}, handling skewed workloads~\cite{yang2012towards}, reducing communication overheads~\cite{gill2018study},
and scalability in network bound applications~\cite{ching2015one}. 
Methods that adaptively determine partitioning strategies at run time~\cite{fan2021graphscope}
or are application-driven~\cite{fan2022application} 
also motivate the need for our solution that employs considerations specific to GCNs during partitioning stage.

\subsection{Distributed Systems for GNNs}

Graph learning tasks perform both forward and backward propagation of model parameters, involving $k$-hop neighborhood aggregations, which require different considerations in partitioning compared to that of traditional graph processing. 
To efficiently train GCNs, methods have been devised to restrict the neighborhood considered by sampling, pruning, and caching~\cite{chen2018fastgcn, zhou2021accelerating, miao2021het}.

Memory management and distributed training are essential for scalable deep neural networks~\cite{demirci2021partitioning, wang2016database,band2020memflow,zhang2021Cerebro}. 
Various frameworks use distributed memory systems for parallel Graph Neural Network (GNN) training~\cite{tgl2022, parallel2021}.
On GPUs, NeuGraph~\cite{ma2019neugraph} uses dataflow scheduling
while ROC~\cite{jia2020improving} utilizes dynamic regression-based partitioning to optimize communication, 
and G3~\cite{liu2020g3} 
leverages graph native operations. 
To reduce communication in full-batch GNN training, 
CAGNET~\cite{tripathy2020reducing} uses the aggregate memory of GPU clusters and the NCCL multi-GPU communication library. 
The DGCL~\cite{cai2021dgcl} communication library instead reroutes communications to use fast links with vertex replication.
Despite distributing the graph data, none of these solutions adopt locality-aware partitioning to reduce communication overheads without replication as we do.

As an alternative to whole-graph training systems, sampling approaches 
overcome the coordination and communication overheads through mini-batches~\cite{serafini2021scalable}. 
In DistDGL~\cite{zheng2020distdgl}, reduction of network communication traffic is achieved by partitioning and co-locating the vertex/edge features with their corresponding local partition data 
for distributed CPU. 
Solutions like AliGraph~\cite{yang2019aligraph}, AGL~\cite{zhang2020agl}, and PaGraph~\cite{lin2020pagraph} all optimize the sampling step in different ways.
We also make use of mini-batch sampling techniques. By integrating the sampling step into our stochastic hypergraph construction phase, we reflect the randomness in communication volumes more accurately. 
DistDGLv2~\cite{zheng2022distributed} recently achieves a hybrid design 
with asynchronous sampling to overlap CPU and GPU computations, which motivates a future blended approach with our CPU/GPU versions of our communication scheme.

The resource under-utilization problem is worse in GPUs since mini-batch sampling overshadows training time~\cite{serafini2021scalable}, especially for sparse graphs~\cite{zheng2022bytegnn}. 
Hence, many works, including our own, 
focus on CPU implementations.
The communication architectures on CPU also involve different optimization considerations compared to GPU-based systems~\cite{min2021large}.
For example, Dorylus explores a CPU-based serverless asynchronous pipeline for scalability~\cite{thorpe2021dorylus}.
ByteGNN~\cite{zheng2022bytegnn} recently 
improves resource utilization on CPUs 
with a partitioning strategy tailored for GNN sampling, however does not account for sparsity as we do, which gives us better speedups. 


\subsection{Data-Parallelization for GNNs}
There are numerous studies on improving the efficiency of GNN computations, such as in cloud data processing systems on top of MapReduce~\cite{ghoting2011systemml} or Hadoop~\cite{huang2013cumulon}.
Ours is a data-parallel approach designed specifically for distributed training of GNNs, utilizing non-blocking parallel SpMM alongside local DMM. 
We make use of sparsity-based partitioning guided by a hypergraph model, to achieve non-blocking point-to-point communications for lowering communication costs.
The potential of exploiting such data access patterns in GNN training has been recently highlighted as an open research question~\cite{kumar2021Cerebro}. 
Graph convolution computations and GCN training depend highly on SpMM and DMM operations, therefore considering the access patterns in these computations is important in improving the resource efficiency and scalability of GCN training. The GE-SpMM algorithm~\cite{huang2020gespmm} for GPUs allows integration with DGL for faster computation of GNNs by processing columns in parallel and ensuring coalesced access to sparse matrix data. Feat-Graph~\cite{hu2020featgraph} co-optimizes graph traversal and feature dimension computation to offer efficient CPU/GPU implementations of sampled dense-dense matrix product~(SDDMM) and SpMM in GNN training. FusedMM develops a general-purpose matrix multiplication kernel for graph embedding and GNN operations~\cite{rahman2021fusedmm}. FusedMM unifies the matrix multiplications into a single operation since SpMM is frequently directly followed by DMM, but the approach is only applicable to shared-memory systems.
In our algorithm, beyond point-to-point communications for SpMM and data locality for DMM, we pay special attention to the requirements of forward propagation and backpropagation during the training phase (aggregating features/gradients and updating parameters), and introduce a stochastic method to handle mini-batch sampling.

\section{Background}
\label{sec:back}

\subsection{Graph Convolution}\label{sec-PGCN}
\label{sec:conv}


GCNs generalize the convolution operation to graphs having arbitrary size and topology, using an adjacency matrix to describe the (sparse) edge connections along which data aggregation takes place for every layer in the neural network. 

Let $\mA\!\in\!\mathbb{R}^{n\times n}$ denote the adjacency matrix of a graph $\Gg\!=\!(\Vs,\Es)$ which consists of $|\Vs|\!=\!n$ vertices.
Vertex set $\Vs$ is associated with a feature matrix $\mH^{k}\!\in\!\mathbb{R}^{n\times d_k}$ for every GCN layer, rows of which correspond to $d_k$-dimensional vertex features.
Given an input feature matrix $\mH^0$, feedforward of GCN is defined as 
\begin{align}
\mZ^{k}&=\widehat{\mA} \mH^{k-1} \mW^{k} \nonumber \\
\mH^{k}&=\sigma(\mZ^{k}) 
\label{equ:feedforward}
\end{align}
\noindent for layers $k\!=\!1,2,\ldots L$. Matrix $\widehat{\mA}$ is formed as $\widehat{\mA}\!=\!\mD^{-\frac{1}{2}}\widetilde{\mA}\mD^{-\frac{1}{2}}$ for normalization, where matrix $\widetilde{\mA}\!=\!\mA\!+\!\mat{I}$ corresponds to the adjacency matrix with self loops and matrix $\mD(i,i)\!=\!\sum_{j}\widetilde{\mA}(i,j)$ corresponds to the diagonal matrix of vertex degrees.
To ease the notation, we will use $\mA$ instead of $\widehat{\mA}$ to denote the normalized adjacency matrix.
In Equation~(\ref{equ:feedforward}), only the $\mA$ matrix is sparse and the remaining matrices are dense.
SpMM $\mA\mH^{k-1}$ combines feature vectors for each vertex (itself and neighbors).
The resulting combined features are then involved in a DMM and multiplied by trainable parameter matrix $\mW^{k}\!\in\!\mathbb{R}^{d_{k-1}\times d_{k}}$.
Finally, a non-linear activation function $\sigma(\cdot)$ is applied to each element of matrix $\mZ^{k}$ to compute $\mH^{k}$.

The backpropagation phase requires a gradient matrix $\mG^{L}\!\in\!\mathbb{R}^{n\times d_{L}}$ which is computed as
\begin{equation}
\mG^{L} = \nabla_{\mH^{L}} \textbf{J} \ \odot \sigma^{\prime}(\mZ^{L})
\end{equation}
\noindent 
where $\nabla_{\mH^{L}}\textbf{J}$ denotes the matrix of derivatives of the loss function $\textbf{J}$ with respect to output features in $\mH^{L}$, $\sigma^{\prime}(\cdot)$ denotes the derivative of the activation function, and symbol $\odot$ denotes element-wise multiplication~(i.e., Hadamard product).  
Gradient matrices for the preceding layers for $k\!=\!L,L\!-\!1,\ldots,1$ are recursively computed as 
\begin{align}
\mS^{k}&=\mA\mG^{k}(\mW^{k})^{T}\nonumber \\
\mG^{k-1}&=\mS^{k}\odot\sigma^{\prime}(\mZ^{k-1})
\label{equ:backpropagation}
\end{align}
In Equation~(\ref{equ:backpropagation}), SpMM is performed with matrices $\mA$ and $\mG^{k}$, and the resulting matrix is used in DMM with $(\mW^{k})^{T}$. 
Each gradient matrix $\mG^{k}\!\in\!\mathbb{R}^{n\times d_{k}}$ is used to update parameter matrix $\mW^{k}$ by the following set of gradient update rules
\begin{align}
\Delta\mW^{k} &= (\mH^{k-1})^{T}\mA\mG^{k} \\
\mW^{k} &\leftarrow \mW^{k} - \eta \Delta\mW^{k}
\end{align}
\noindent where $\Delta\textbf{W}^{k}$ denotes the matrix of derivatives of the loss function $\textbf{J}$ with respect to parameters in matrix $\mW^{k}$, and $\eta$ denotes the learning rate. It is important to note that, if the input graph is directed, transpose $\mA^T$ is used instead of $\mA$ in backpropagation~(we refer the reader to~\cite{tripathy2020reducing} for a more detailed description).


\subsection{Graph and Hypergraph Partitioning}
\label{sec:hgp}
Given a graph $\Gg\!=\!(\Vs,\Es)$ with vertex set $\Vs$ and edge set $\Es$, a $p$-way partitioning of $\Gg$ is defined as $\Pi\!=\!\{\Vs_1, \Vs_2 \cdots \Vs_p\}$ consisting of subsets of vertices $\Vs_m\!\subset\!\Vs$ that are mutually disjoint ($\Vs_m\cap \Vs_n\!=\!\emptyset$ if $m\!\neq\!n$) and nonempty ($\Vs_m\!\neq\!\emptyset$ $\forall \Vs_m\!\in\!\Pi$) where union of these subsets gives the vertex set~($\bigcup \Vs_m = \Vs)$. 

Each undirected edge $\{v_i, v_j\}\!\in\!\Es$ between vertices $v_i, v_j\!\in\!\Vs$ is given a $\mathrm{cost}(v_i, v_j)$ and each vertex $v_i\!\in\!\Vs$ is associated with a weight $w(v_i)$, therefore the weight of a part $\Vs_m\!\in\!\Pi$ is defined as $W(\Vs_m)\!=\!\sum_{v_i \in \Vs_m}w(v_i)$.
The partition $\Pi$ is balanced if it satisfies the balancing constraint $W(V_m) \leq W_{avg} (1 + \epsilon)$ for all $V_m\!\in\! \Pi$ where ${W_{avg} = \sum_{v_i \in V}w(v_i) / p}$ is the average part weight and $\epsilon$ is the maximum allowed imbalance ratio. 
Under a partitioning $\Pi$, an undirected edge $\{v_i, v_j\}\!\in\!\Es$  is called cut edge if it connects vertices belonging two different parts.
The $p$-way graph partitioning problem is defined as finding a partitioning $\Pi$ such that the balancing constraint is satisfied and the total partitioning cost $\chi(\Pi) = \sum_{ \{v_i, v_j\} \in \Es_C} \mathrm{cost}(v_i, v_j)$ is minimized where $\Es_C$ denotes the set of cut edges.

Hypergraphs generalizes graphs by allowing hyperedges~(nets) to connect more than two vertices. Let $\Hg\!=\!(\Vs,\Ns)$ denote a hypergraph consisting of vertex set $\Vs$ and net set $\Ns$, with $\Pi$ defined as above. The set of vertices connected by a net $n_j\!\in\!\Ns$ is denoted by $\mathrm{pins}(n_j)$, where each net $n_j$ is associated with $\mathrm{cost}(n_j)$.
Under the partition $\Pi$, the connectivity set $\Lambda(n_j)$ is the set of parts that net $n_j$ connects~(i.e., $\mathrm{pins}(n_j) \cap \Vs_m\!\neq\!\emptyset$). The number of such parts is called connectivity $\lambda(n_j)\!=\!|\Lambda(n_j)|$.
If a net $n_j$ connects to multiple parts~(i.e., $\lambda(n_j)\!>\!1$) it is said to be cut, and uncut otherwise.
The connectivity cut size under $\Pi$ is defined as $\chi(\Pi) = \sum_{n_j \in \Ns} \mathrm{cost}(n_j)\times(\lambda(n_j) -1)$.
The hypergraph partitioning problem is therefore finding a $p$-way partition that satisfies the balancing constraint while minimizing the cut size, and is NP-Hard. 
There are successful tools that produce quality results for both graph and hypergraph partitioning problems~\cite{catalyurek1999hypergraph,karypis1998hmetis}.

\begin{figure*}[t]
     \centering
     \begin{subfigure}[b]{\textwidth}
         \centering
    		 \includegraphics[width=0.77\textwidth]{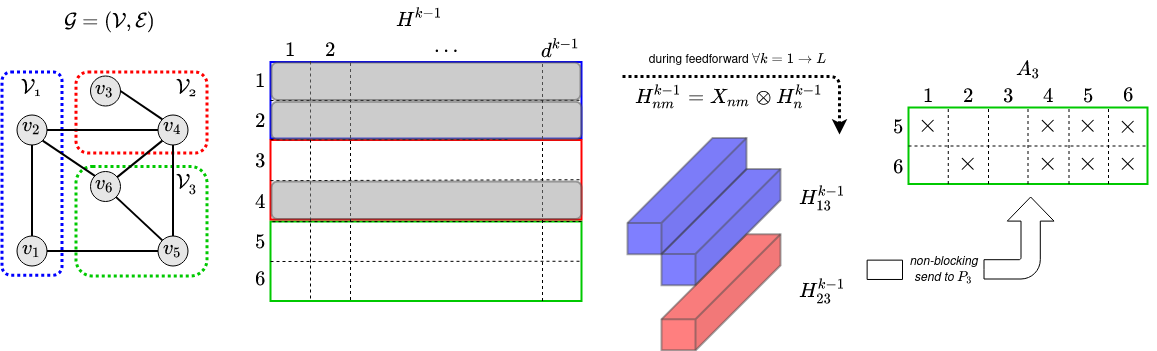}
         \caption{For graph $\Gg$, the feature matrix $\mH^{k-1}$ for layer $k-1$ has been conformably partitioned along with $\mA$ while the weight matrix $\mW^k$ has been duplicated across all $3$ processors. Processor $P_3$ (in green) requires $\mH^{k-1}_{13}$ and $\mH^{k-1}_{23}$ from the other two processors (in blue and red respectively).}
         \label{fig:1-1}
     \end{subfigure}
     \hfill
     \begin{subfigure}[b]{\textwidth}
         \centering
    		 \includegraphics[width=0.77\textwidth]{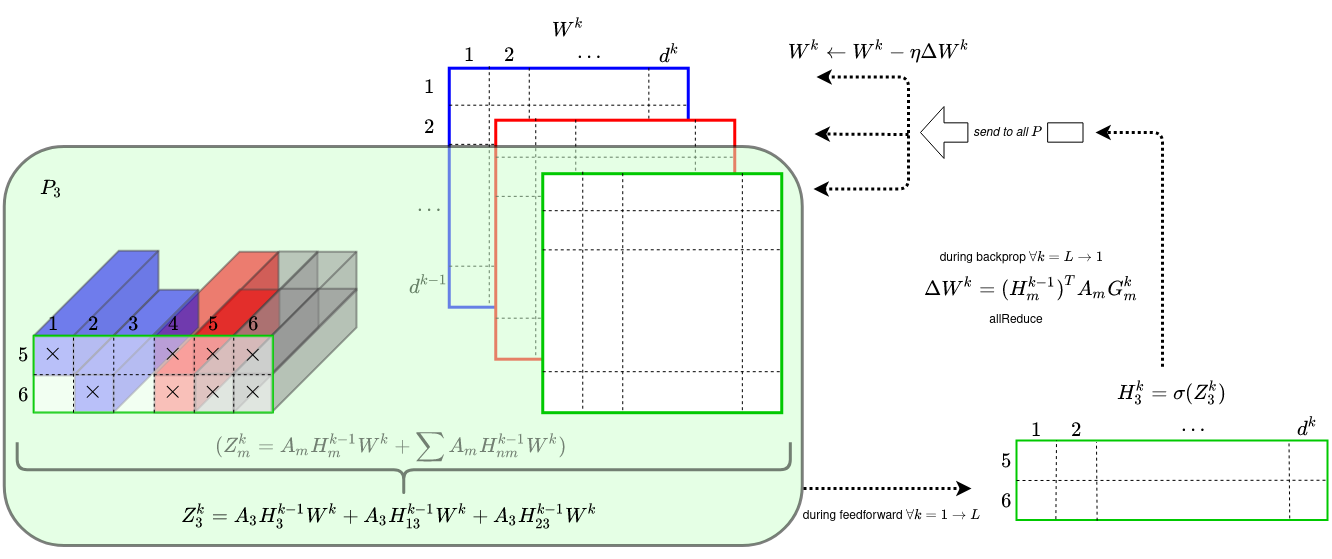}
         \caption{Computations are demonstrated here on the processor $P_3$ (in green) which stores $\mA_3$ and $\mH^{k-1}_3$ locally but must receive $\mH^{k-1}_{13}$ and $\mH^{k-1}_{23}$. The resulting $\mH^k_3$ features are similarly computed for all layers ($k=1 \rightarrow L$) during the feedforward phase. Subsequently, during backpropagation ($k=L \rightarrow 1$), these are used to generate $\Delta\mW^k$ for updation of the weight matrices $\mW^k$ on all processors.}
         \label{fig:1-2}
     \end{subfigure}
     \label{fig:1}
     \caption{Communication and computation processes in feedforward and backpropagation of the distributed GCN algorithm.}
     \label{fig:1}
\end{figure*}

\section{Parallel GCN Training}
\label{sec:method}

We first present the feedforward and backpropagation steps of the proposed algorithm. Next, we describe a hypergraph partitioning model which reduces communication overheads over the graph model. 
We also propose a stochastic hypergraph model which encodes expected communication volume in mini-batch training.

\subsection{Feedforward}
\label{sec:feedforward}

The proposed parallel feedforward algorithm executes on $p$ processors each of which is denoted by $P_m$ for $m\!=\!1,2,\ldots p$.
Adjacency matrix $\mA$ and vertex feature matrices $\mH^{k}$ for all layers $k\!=\!0,1,\ldots, L$ are 1D row-wise partitioned among processors where each processor $P_m$ stores submatrices $\mA_m\!\in\!\mathbb{R}^{n\times n}$ and $\mH^{k}_m\!\in\!\mathbb{R}^{n\times d_k}$,
which only contain subsets of rows of matrices $\mA$ and $\mH^{k}$.
Adjacency matrix and feature matrices are conformably partitioned so that if row $\mA(i,:)$ is assigned to submatrix $\mA_m$, then the corresponding feature vectors $\mH^k(i,:)$ for all layers $k$ are assigned to submatrices $\mH^k_m$, respectively~(i.e., $\mA(i,:)\!\in\!\mA_m \Leftrightarrow \mH^k(i,:)\!\in\!\mH^k_m \, \forall k$).
Parameter matrices $\mW^{k}$ for all layers $k$ are replicated and stored by all processors due to their relatively smaller sizes.

The matrix partitioning scheme encodes a vertex-partitioning on graph $\Gg$, since rows $\mA(i,:)$ and $\mH^{k}(i,:)$ denote the adjacency list and features of vertex $v_i\!\in\!\Vs$.
Moreover, this partitioning also induces a task partitioning in feedforward phase:
If a vertex $v_i$ is assigned to a processor $P_m$, the task of computing row $\mZ(i,:)^{k}$ of intermediate matrix $\mZ^{k}$ in layer $k$ is performed by processor $P_m$ where row $\mZ(i,:)^{k}$ is computed as
\begin{equation}
\mZ^{k}(i,:) = \left( \sum\limits_{j \in \mathrm{cols}(\mA(i,:))}\mA(i,j) \mH^{k-1}(j,:) \right)  \mW^{k}. \label{eq:atomic-task1}
\end{equation}
\noindent
Hence, to compute submatrix $\mZ^{k}_m$, processor $P_m$ needs to receive all $\mH^{k-1}$-matrix rows corresponding to all nonzero column indices in $\mA_m$, which are not locally stored in $\mH^{k-1}_m$.
Let $\mH^{k-1}_{nm}\!\in\!\mathbb{R}^{n\times d_{k-1}}$ denote the submatrix consisting of rows that are needed to be transferred from processor $P_n$ to $P_m$. 
That is, submatrix $\mH^{k-1}_{nm}$ contains subset of rows of $\mH^{k-1}_n$ corresponding to the intersection of nonzero row indices of $\mH^{k-1}_n$ and column indices of $\mA_m$. 
More formally, $\exists i\!\in\!\mathrm{rows}(\mH^{k-1}_{nm})$ if $i\!\in\!\mathrm{cols}(\mA_{m})\!\cap\!\mathrm{rows}(\mA_{n})$.
We use non-blocking point-to-point communications to transfer these submatrices between processors.
After processor $P_m$ receives submatrix $\mH^{k-1}_{nm}$ from each processor $P_n$ for all $n\!\neq\!m$ such that $\mH^{k-1}_{nm}\!\neq\!\mathbf{0}$, $P_m$ performs multiplication
\begin{equation}
\mZ^{k}_m\!=\! (\mA_m\mH^{k-1}_m  + \sum\limits_{n\neq m}\mA_m\mH^{k-1}_{nm})\mW^{k}
\end{equation}
\noindent
to compute submatrix $\mZ^{k}_m\!\in\!\mathbb{R}^{n\times d_{k}}$.
Then, $P_m$ applies the nonlinear activation function \mbox{$\mH^{k}_m\!=\!\sigma(\mZ^{k}_m)$} to proceed to the next layer.

To manage sparse point-to-point communication operations, each processor $P_m$ is provided with sets $\Ss_m$ and $\Rs_m$ which are computed before training with respect to the partitioning of adjacency matrix $\mA$ among processors.
Set $\Ss_m$ is composed of diagonal matrices $\mX_{mn}\!\in\!\mathbb{R}^{n\times n}$ for each processor $P_n\!\neq\!P_m$.
Matrix $\mX_{mn}$ is used in a special matrix multiplication to determine which local $\mH^{k-1}_m$-rows to be sent by processor $P_m$ to $P_n$.
Formally, 
\begin{equation}
\begin{split}
\Ss_m = \left\lbrace \mX_{mn} \mid \mX_{mn}\neq \mathbf{0} \wedge \mX_{mn}(i,i) = 1 \;\; \right. \\ \left. \forall i \in \mathrm{cols}(\mA_{n}) \cap \mathrm{rows}(\mA_{m}) \right\rbrace.
\end{split}
\end{equation}
\noindent
That is, the $i$th diagonal entry $\mX_{mn}(i,i)\!=\!1$ if the intersection of nonzero row and column indices of matrices $\mA_m$ and $\mA_n$ contains index $i$, otherwise it is set to zero.
Set $\Rs_m$ is composed of processors from which $P_m$ receives at least one message.
Formally,
\begin{equation}
\begin{split}
\Rs_m = \left\lbrace P_n \mid  P_n \neq P_m  \wedge cols(\mA_{m}) \cap rows(\mA_{n}) \neq \emptyset \right\rbrace 
\end{split}.
\end{equation}
\noindent
That is, processor $P_n$ is included in $\Rs_m$ if the intersection of nonzero row and column indices of matrices $\mA_m$ and $\mA_n$ is nonempty, and processor $P_m$ receives at least one row of $\mH^{k-1}_n$ from $P_n$.

Algorithm~\ref{algorithm:feedforward} describes the proposed parallel feedforward algorithm.
We used SuiteSparse:GraphBLAS~(GB)~\cite{davis2019algorithm} library to perform the sparse matrix operations.
In lines~$3$--$5$, to overlap communication by computation, a non-blocking communication is performed  for each diagonal matrix $\mX_{mn}\!\in\!\Ss_m$ by processor $P_m$ to send required $\mH^{k-1}_m$-matrix rows to processor $P_n$.
Matrix $\mH^{k-1}_{mn}$ is formed through a specialized matrix multiplication \mbox{$\mH^{k-1}_{mn}\!=\!\mX_{mn}\!\otimes \mH^{k-1}_{m}$}.
By this matrix multiplication, if the $i$th diagonal entry is $\mX_{mn}(i,i)\!=\!1$, then the $i$th row $\mH^{k-1}_m$ is copied into matrix $\mH^{k-1}_{mn}$.
Operator $\otimes$ denotes that the matrix multiplication is performed under semiring $\mathrm{GxB\_PLUS\_SECOND}$, defined by GB library, to replace the multiplication operator with a copy operator that will directly carry the second operand to the resulting variable without multiplying~(i.e., $z\!=\!x\times y \Rightarrow z\!=\!y$).
In line~$6$, local matrix multiplication $\mZ^{k}_m\!=\!\mA_{m}\mH^{k-1}_{m}\mW^{k}$ is performed without waiting for the non-blocking communication operations to complete.
Matrix $\mZ^{k}_m$ is incomplete at this stage and its computation is finalized after receiving all necessary data.
In lines~$7$--$9$, processor $P_m$ receives $\mH^{k-1}_{nm}$ from each $P_n\!\in\!\Rs_m$, and performs multiplication and addition $\mZ^{k}_m = \mZ^{k}_m + \mA_{m} \mH^{k-1}_{nm} \mW^{k}$ to compute the final matrix $\mZ^{k}_m$.


\begin{algorithm}[t]
\SetKwInOut{Input}{input}
\SetAlgoLined
\ForAll{\emph{processors} $P_m$ \textbf{\emph{in parallel}}}{
 \For{$k=1$ \textbf{\emph{to}} $L$} {
  \ForEach{$\mX_{mn} \in \Ss_m$ }{
   $\mH^{k-1}_{mn} = \mX_{mn} \otimes \mH^{k-1}_{m}$\\
   Non-blocking send $\mH^{k-1}_{mn}$ to processor $P_n$\\
  }
  $\mZ^{k}_m = \mA_{m} \mH^{k-1}_{m} \mW^{k}$\\
  \ForEach{$P_n \in \Rs_m$ }{
   Receive $\mH^{k-1}_{nm}$ from processor $P_n$\\
   $\mZ^{k}_m = \mZ^{k}_m + \mA_{m} \mH^{k-1}_{nm} \mW^{k}$
  }
  $\mH^{k}_{m} = \sigma(\mZ^{k}_m)$\\
 }
}
\caption{Parallel Feedforward}
\label{algorithm:feedforward}
\end{algorithm}

Figure~\ref{fig:1} displays a sample execution of the feedforward phase.
The adjacency matrix $\mA$, and feature matrix $\mH^k$ for each layer $k$ are conformably partitioned among the three processors. Thus, each processor $P_m$ only stores submatrices $\mA_m$ and $\mH^{k-1}_m$.
For instance, the computation of matrix $\mZ^k_3$ by processor $P_3$ (in green) requires the other two processors to send $\mH^{k-1}_{13}$ and $\mH^{k-1}_{23}$ corresponding to nonzero indices, and local matrix multiplication is performed without waiting for the completion of these non-blocking communications, to compute $\mH^k_3$. 
Processor $P_3$ retrieves features of $v_1$ and $v_4$ for convolution on vertex $v_5$, and features of $v_2$, $v_4$ for vertex $v_6$.
Hence, $\mH^{k-1}_{13}$ contains rows $1$ and $2$ of $\mH^{k-1}$ while $\mH^{k-1}_{23}$ contains row $4$, since these are the nonzero indices of $\mA_m$ where the feature matrix rows are not locally stored. Note that row $4$ is only transferred once to avoid a redundant communication.

\subsection{Backpropagation}
\label{sec:backprop}

In the backpropagation phase, similar to the vertex feature matrices, gradient matrices $\mG^{k}$ for each layer $k$ are row-wise partitioned among processors where each processor $P_m$ holds submatrix $\mG^{k}_m\!\in\!\mathbb{R}^{n\times d_{k}}$ in each layer $k$.
Gradient matrix $\mG^{k}$ and adjacency matrix $\mA$ are conformably partitioned so that if row $\mA(i,:)$ is assigned to submatrix $\mA_m$, then row $\mG^k(i,:)$ is assigned to submatrix $\mG^k_m$~(i.e., $\mA(i,:)\!\in\!\mA_m \Leftrightarrow \mG^k(i,:)\!\in\!\mG^k_m \, \forall k$).
Hence, the task of computing row $\mS(i,:)^{k}$ of intermediate matrix $\mS^{k}$ is given to processor $P_m$ if row $\mA(i,:)$ and corresponding vertex $v_i$ is assigned to $P_m$.
So, the same row-wise partitioning is induced on matrix $\mS^{k}$ as with matrices $\mA$ and $\mG^{k}$.
Matrix $\mS^{k}$ is computed by following similar steps of computation of $\mZ^{k}$ in feedforward phase.
Then, $P_m$ performs element-wise multiplication $\mG^{k-1}_m\!=\!\mS^{k}_m\odot\sigma^{\prime}(\mZ^{k-1}_m)$.

\begin{algorithm}[t]
\SetKwInOut{Input}{input}
\SetAlgoLined
\ForAll{\emph{processors} $P_m$ \textbf{\emph{in parallel}}}{
 $\mG^{L}_m = \nabla_{\mH^{L}_m} \textbf{J} \ \odot \sigma^{\prime}(\mZ^{L}_m)$\\
  \For{$k=L$ \textbf{\emph{to}} $1$} {
   \ForEach{$\mX_{mn} \in \Ss_m$ }{
    $\mG^{k}_{mn} = \mX_{mn} \otimes \mG^{k}_{m}$\\
    Non-blocking send $\mG^{k}_{mn}$ to processor $P_n$\\
   }
   $\mS^{k}_m = \mA_{m} \mG^{k}_{m} (\mW^{k})^{T}$\\
  \ForEach{$P_n \in \Rs_m$ }{
    Receive $\mG^{k}_{nm}$ from processor $P_n$\\
    $\mS^{k}_m = \mS^{k}_m + \mA_{m} \mG^{k}_{nm} (\mW^{k})^{T}$
   }
   $\mG^{k-1}_m=\mS^{k}_m\odot\sigma^{\prime}(\mZ^{k-1}_m)$\\
   $\Delta\textbf{W}^{k}_m = (\mH^{k-1}_m)^{T}(\mA_m\mG^{k})$\\
   $\Delta\textbf{W}^{k}$ = $\operatorname{Allreduce-sum}(\Delta\textbf{W}^{k}_m)$ \\ 
   $\textbf{W}^{k} \leftarrow \textbf{W}^{k} - \eta \Delta\textbf{W}^{k}$\\
  }
}
\caption{Parallel Backpropagation}
\label{algorithm:backpropagation}
\end{algorithm}

Algorithm~\ref{algorithm:backpropagation} gives the proposed parallel backpropagation algorithm.
In line~$2$, each processor $P_m$ computes submatrix $\mG^{L}_m$ by using the local vertex-feature matrix $\mH^{L}_m$ in the final layer.
Here, $\nabla_{\mH^{L}_m} \textbf{J}$ denotes the matrix of partial derivatives of the loss function with respect to $\mH^{L}_m$, and its formulation depends on the definition of the loss function.
In lines~$4$--$10$, matrix $\mS^{k}$ is computed in a similar way to computation of $\mZ^{k}$ in Algorithm~\ref{algorithm:feedforward}. 
In line~$11$, gradient matrix $\mG^{k-1}$ for the preceding layer is computed via element-wise multiplication of matrices $\mS^{k}_m$ and $\sigma^{\prime}(\mZ^{k-1}_m)$.
In line~$12$, each processor $P_m$ computes partial results for gradient matrix $\Delta\textbf{W}^{k}$ of the loss function $\textbf{J}$ with respect to parameter matrix $\mW^{k}$.

In the computation of $\Delta\textbf{W}^{k}$, matrix $(\mH^{k-1}_m)^{T}$ is computed in feedforward phase, whereas $(\mA_m\mG^{k})$ part is computed as a by-product in lines~$7$ and~$10$.
Here, if column $(\mH^{k-1})^{T}(:,i)$ is stored in $(\mH^{k-1}_m)^{T}$, then the corresponding row $(\mA\mG^{k})(i,:)$ is also stored in $(\mA_m\mG^{k})$.
Therefore, multiplication $(\mH^{k-1}_m)^{T}(\mA_m\mG^{k})$ by processor $P_m$ produces matrix $\Delta\textbf{W}^{k}_m$ of partial products where each nonzero $\Delta\textbf{W}^{k}_m(i,j)$ contributes to the corresponding nonzero
\[\Delta\textbf{W}^{k}(i,j)\!=\!\sum_m \Delta\textbf{W}^{k}_m(i,j)\]
in the final matrix $\Delta\textbf{W}^{k}$.
In line~$13$, the final gradient matrix $\Delta\textbf{W}^{k}$ is computed via an allreduce-type communication operation which combines~(sums) partial matrices from all processes and distributes the result back to all processes.
In line~$14$, gradient update on $\mW^{k}$ is performed by all processors on their local copies.



Figure~\ref{fig:1} also displays the additional computations performed in the backpropagation phase.
As seen in the figure, the relatively smaller-sized weight matrices $\mW^k$ for each layer $k$ are replicated among all processors.
The computation of matrix $\mS^k$ is identical with the computation of matrix $\mH^k$ and requires the same communication steps which are determined by the partitioning on the adjacency matrix $\mA$. Matrix $\mS^k$ is used together with matrix $\mZ^{k-1}$ to compute gradient matrix $\mG^{k-1}$.
The figure also shows the all-reduce operation performed on locally computed matrices $\Delta\mW^k_m$ to compute the final matrix $\Delta\mW^k$ for gradient update operations.

\subsection{Partitioning Models}
\label{sec:part}

Different partitioning models may be used for partitioning the adjacency matrix among processors. 
We compare the graph and hypergraph models and highlight how the hypergraph model correctly encodes the total communication volume during the message-passing operations.
Finally, we present our novel stochastic hypergraph model which encodes expected communication volume instead of exact values, and therefore supports mini-batch training.

\subsubsection{Graph Model}
~\\
In a graph model, a $p$-way partitioning $\Pi_{p}\!=\!\{\Vs_1,\Vs_2,\ldots,\Vs_{p}\}$ over vertex set $\Vs$ induces a row-wise partitioning on matrix $\mA$ among $p$ processors.
If a vertex $v_i$ is assigned to part $\Vs_m\!\in\!\Pi_{p}$, then row $\mA(i,:)$ is assigned to processor $P_m$.
Note that the input graph $\Gg\!=\!(\Vs,\Es)$ in GCN training can be directed or undirected, but graph partitioning tools~(e.g., METIS) assume that the graph to be partitioned is undirected, edges have integer costs, and vertices have integer weights.
Therefore, we build an undirected graph $\Gg^{\prime}\!=\!(\Vs,\Es^{\prime})$ where we use vertex set $\Vs$ as is, but replace each directed edge $(v_i,v_j)\!\in\!\Es$ with an undirected edge $\{v_i,v_j\}\!\in\!\Es^{\prime}$.

Under a partition $\Pi_p$, each undirected cut edge $\{v_i,v_j\}$ represents the communication of $\mH^{k-1}(i,:)$- and $\mH^{k-1}(j,:)$-matrix rows between respective processors during feedforward phase, and communication of $\mG^{k}(i,:)$- and $\mG^{k}(j,:)$-matrix rows during backpropagation phase.
Since we have $d$-dimensional vertex feature matrix $\mH^{k}\!\in\!\mathbb{R}^{n\times d_k}$, each undirected edge encodes a total communication volume of $\sum_k 2(d_{k-1}\!+\!d_k)$ nonzero entries over all layers $k$.
Because the communication volume encoded by each edge is the same constant value, each undirected edge $\{v_i,v_j\}\!\in\!\Es^{\prime}$ can be associated with a unit $\mathrm{cost}(v_i, v_j)\!=\!1$.
Each vertex $v_i$ is associated with a computational weight $w(v_i)\!=\!|\mathrm{cols}(\mA(i,:))|$.
DistDGL~\cite{zheng2020distdgl} also utilizes this partitioning scheme and partitions the input graph via METIS, only considering undirected graphs.

What makes the graph model less accurate compared to the hypergraph model is that the former overestimates the total communication volume between processors.
This deficiency of the graph model can be seen in two ways:
(i)~When both of the directed edges $(v_i,v_j)$ and $(v_j,v_i)$ are not simultaneously present in the input graph $\Gg$, the graph model still considers an undirected edge $\{v_i,v_j\}$ that sees communication in both ways although the communication is actually one-way.
(ii)~If a vertex $v_i$ is connected to vertices $v_j$ and $v_k$ that are stored together but on a different processor from $v_i$, the graph model assumes that the features of $v_i$ are sent twice. However, these features are sent to that processor once in a single message.
These two cases cause the partitioning cut size to be higher than the actual communication volume.

\subsubsection{Hypergraph Model}
~\\
We model one-dimensional~(1D) row-wise partitioning of adjacency matrix as a hypergraph partitioning problem~\cite{catalyurek1999hypergraph} since the hypergraph model can encode the exact communication volume of parallel GCN. The connectivity cut size of the hypergraph model encodes the total communication volume among processors, while weights of partitions encode the associated computational load for processors. 
Hence, minimization of the connectivity cut size under weight-balancing constraints achieves minimization of the total communication volume while achieving computational-load balance. 
During the feedforward phase, the hypergraph model encodes the total communication volume on $\mH^{k-1}$-matrix rows for parallel SpMMs $\mA\mH^{k-1}$ among processors in each layer $k$. The model also encodes the total communication volume on $\mG^{k}$-matrix rows for parallel SpMMs $\mA\mG^{k}$ during backpropagation phase. 

\begin{figure*}
    \centering
    \includegraphics[width=\textwidth]{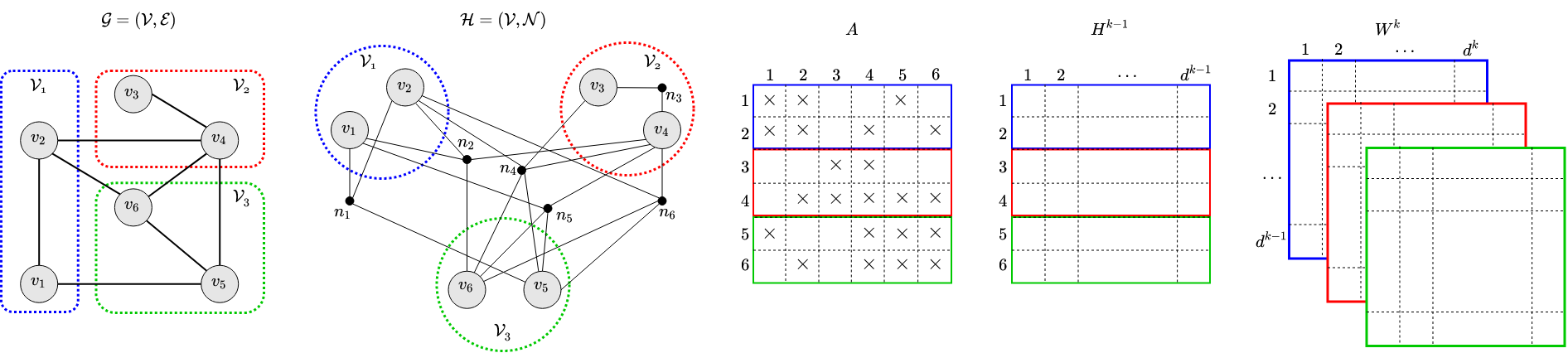}
    \caption{Hypergraph partitioning of graph $\Gg$ having adjacency matrix $\mA$ (including self loops), by constructing the corresponding hypergraph $\Hg$ where every net $n_j$ connects nonzero entries of the column $i$ in $\mA$. The feature matrix $\mH^k$ for layer $k$ is conformably partitioned along with $\mA$, while the weight matrix $\mW^k$ is duplicated across all processors.}
    \label{fig:2}
\end{figure*}

To partition adjacency matrix $\mA$, we first build a hypergraph $\Hg\!=\!(\Vs,\Ns)$ where for each matrix row $\mA(i,:)$ there exists one vertex $v_i\!\in\!\Vs$ and for each column $\mA(:,j)$, there exists one net $n_j\!\in\!\Ns$.
Similar to the graph model, a partitioning obtained on the vertex set of the input graph $\Hg\!=\!(\Vs,\Es)$ also induces a 1D row-wise partitioning on the adjacency matrix.
That is, a $p$-way partitioning $\Pi_{p}\!=\!\{\Vs_1,\Vs_2,\ldots,\Vs_{p}\}$ over vertex set $\Vs$ induces a row-wise partitioning on matrix $\mA$ among $p$ processors, since each vertex $v_i$ corresponds to row $\mA(i,:)$.
Additionally, each vertex $v_i\in\!\Vs$ also represents the task of computing rows $\mZ^{k}(i,:)$ and $\mS^{k}(i,:)$ in each layer $k$.
Therefore, each vertex $v_i$ is associated with weight $w(v_i)\!=\!|\mathrm{cols}(\mA(i,:))|$, i.e., the number of nonzero column indices in the $i$th row of matrix $\mA$, to encode the computational load of the task represented by vertex $v_i$.
Note that the number of nonzero arithmetic operations required to compute rows $\mZ^{k}(i,:)$ and $\mS^{k}(i,:)$ is proportional to the number of nonzero column indices in row $\mA(i,:)$.
So, satisfying the balancing constraints in hypergraph partitioning achieves computational-load balance.

Net set $\Ns$ encodes task dependencies on rows of matrices $\mH^{k-1}$ and $\mG^{k}$ during feedforward and backpropagation phases for each layer $k$.
Each net $n_j\in\Ns$ connects all vertices $v_i\!\in\!\Vs$ for which the corresponding row $\mA(i,:)$ has a nonzero entry in the $j$th column.
For computing rows $\mZ(i,:)^{k}$ and $\mS(i,:)^{k}$, the processor that owns row $\mA(i,:)$ needs all $\mH^{k-1}$- and $\mG^{k}$-matrix rows, corresponding to nonzero column indices $\mathrm{cols}(\mA(i,:))$, respectively.
Therefore, pins of a net $n_j$ denotes the tasks that require row $\mH^{k-1}(j,:)$ and $\mG^{k}(j,:)$.
Formally, pins of a net $n_j$ can be written as 
\begin{equation}
\mathrm{pins}(n_j)= \{v_i \in V \mid \exists j \in \mathrm{cols}(\mA(i,:))\}.
\end{equation}

Under a partitioning $\Pi_{p}$, a net $n_j\!\in\!\Ns$ with connectivity set $\Lambda(n_j)$ encodes the total communication volume on rows $\mH^{k-1}(j,:)$ and $\mG^{k}(j,:)$ in each layer $k$.
Here, at least one part in $\Lambda(n_j)$ stores vertex $v_j$ since each diagonal entry contains a nonzero entry in adjacency matrix $\mA$.
That is, for all net $n_j\!\in\Ns$, vertex $v_j\!\in\!\mathrm{pins}(n_j)$.
Therefore, a part $V_m\in\Lambda(n_j)$ stores vertex $v_j$ and hence, processor $P_m$ stores rows $\mH^{k-1}(j,:)$ and $\mG^{k}(j,:)$ in its local submatrices $\mH^{k-1}_m$ and $\mG^{k}_m$, respectively.
Due to the task dependencies encoded by net $n_j$, processor $P_m$ sends row $\mH^{k-1}(j,:)$ to all processors corresponding to parts in $\Lambda(n_j)\!\setminus\!\Vs_m$ during feedforward phase, i.e., $\lambda(n_j)-1$ communications.
Similarly, processor $P_m$ sends row $\mG^{k}(j,:)$ to all processors in $\Lambda(n_j)\!\setminus\!\Vs_m$ during backpropagation phase.
If a processor $P_n\in\Lambda(n_j)\!\setminus\!\Vs_m$ has multiple vertices connecting to net $n_j$, processor $P_n$ receives row $\mH^{k-1}(j,:)$ and row $\mG^{k}(j,:)$ only once.
So, net $n_j$ incurs a communication volume of $\mathrm{cost}(n_j)\!\times\!\left(\lambda(n_j)\!-\!1 \right)$ where the cost of net $n_j$ is denoted as \mbox{$\mathrm{cost}(n_j)\!=\!\sum_k d_{k-1}\!+\!d_{k}$} since all nonzero entries in rows $\mH^{k-1}(j,:)$ and $\mG^{k}(j,:)$ are communicated in each layer $k$.
Since the cost of each net is the same constant value, we can also associate each net with a unit cost $cost(n_j)\!=\!1$.
Therefore, the total communication volume can be written as
\begin{equation}
\sum\limits_{n_j \in \Ns} 2 \times \mathrm{cost}(n_j) \times \left( \lambda(n_j)-1 \right)
\end{equation}
\label{eq:comm-vol}
\noindent
which indicates that minimizing the connectivity cut size corresponds to minimizing the total communication volume.

Figure~\ref{fig:2} displays an illustrative example of the proposed hypergraph partitioning model on a sample graph $\Gg$ having adjacency matrix $\mA$. 
The hypergraph $\Hg$ is constructed having parts $\Vs_1$ (blue), $\Vs_2$ (red), and $\Vs_3$ (green) each containing two vertices, with a net $n_j$ for every column $\mA(:,j)$. According to the hypergraph partitioning model, rows of $\mA$ are assigned to processors based on the hypergraph vertex partitioning. For example, row $\mA(i,:)$ will be stored on processor $P_1$ as vertex $v_1$ is assigned to $\Vs_1$. Since $v_1$ represents a task, the computational load is proportional to the number of non-zero columns in row $1$ and is encoded by its weight $w(v_1)\!=\!3$.
Each net connects non-zero entries in a row. For example net $n_2$ connects $pins(n_2) = \{v_1,v_2,v_4,v_6\}$ with connectivity set $\Lambda(n_2)=\{\Vs_1,\Vs_2,\Vs_3\}$. Its connectivity is therefore $\lambda(n_2)\!=\!3$.
The net set $\Ns$ thus encodes task dependencies during the feedforward and backpropagation phases since communication operations on matrices $\mH^{k}$ and $\mG^{k-1}$ are identical and determined by the partitioning on matrix $\mA$.
The feature matrix $\mH^{k-1}$ is conformably partitioned with $\mA$ for each layer of the GCN, while the weight matrix $\mW^k$ is replicated across the three processors.

Figure~\ref{fig:2} also depicts how the graph model overestimates the communication volume.
Features of vertex $v_4$ must be fetched by vertices $v_2$, $v_3$, $v_5$, and $v_6$.
For example, according to the graph model, the feature vector of $v_4$ is encoded as if it were sent from processor $P_2$ to processor $P_3$ twice, but should only be sent once.
Therefore, cut edges connecting to vertex $v_4$ in the graph encodes a communication volume of $3$ instead of the true value of $2$.
On the other hand, the hypergraph model shown on $\Hg$ uses net $n_4$ to encode communications from vertex $v_4$. Since the connectivity of $n_4$ is $\lambda(n_4)\!=\!3$ and hypergraph partitioning minimizes connectivity--1 metric, net $n_4$ encodes the true communication volume as \mbox{$\lambda(n_4)\!-\!1\!=\!2$}.


\subsubsection{Stochastic Hypergraph Model}
~\\
In mini-batch training, a stochastic sampling is applied to the input graph to produce subgraphs on which convolutions are performed.
We propose a novel stochastic hypergraph model which encodes and minimizes the \emph{expected} communication volume in mini-batch training. 
Note that the hypergraph/graph models described earlier encode the communication volume in full-batch training.

We first randomly generate mini-batches~(i.e., subgraphs) using a sampling technique.
Next, for each subgraph, we build a hypergraph that encodes the total communication volume for the mini-batch.
By merging all hypergraphs generated (one per mini-batch), we build a larger hypergraph that can encode the \emph{expected} connectivity of any randomly generated net.
Partitioning the resulting merged stochastic hypergraph minimizes the expected connectivity of a random net, and thus minimizes the expected total communication volume for any randomly generated mini-batch.

More formally, given an input graph $\Gg\!=\!(\Vs,\Es)$, each mini-batch corresponds to a subgraph $\Gg'\!=\!(\Vs'\!\subset\!\Vs, \Es'\!\subset\!\Es)$. 
We generate $b$ mini-batches, each corresponding to a subgraph $\Gg_i'=(\Vs'_i,\Es'_i)$ for $i\!=\!1,2,\ldots,\!b$. 
For each such subgraph $\Gg'$ a hypergraph $\Hg'\!=\!(\Vs',\Ns')$ is built in the same way as in full-batch training.
The stochastic hypergraph $\Hg\!=\!(\Vs\!=\!\bigcup \Vs'_i,\Ns\!=\!\bigcup \Ns'_i)$ is formed by merging all vertex and net sets into the corresponding sets for the merged hypergraph. 
The proposed stochastic hypergraph partitioning process is described in Algorithm~\ref{algorithm:partitioning} which returns a partitioning $\Pi$ of $\Hg$ to determine the row-wise partitioning of the adjacency matrix.

\begin{algorithm}[t]
\SetKwInOut{Input}{input}
\SetAlgoLined
Generate $b$ subgraphs $G_i=(V'_i, E'_i)$ of $G$ for $i=1,2,\ldots,b$\\
Build hypergraph $H_i=(V'_i,N'_i)$ for each $G_i=(V'_i, E'_i)$\\
Build stochastic hypergraph $H=(V=\bigcup\limits_{i=1}^{b} V'_i, N= \bigcup\limits_{i=1}^{b} N'_i)$\\
Partition $p$-way hypergraph $H$ to obtain partitioning $\Pi=\{V_1,V_2,\ldots, V_p\}$\\
\textbf{Return} $\Pi$
\caption{Stochastic Hypergraph Partitioning}
\label{algorithm:partitioning}
\end{algorithm}

Under a $p$-way vertex partition $\Pi$ of the stochastic hypergraph $\Hg$, let $\lambda$ denote the expected connectivity of a randomly generated net.
By using Hoeffding’s inequality, the value of $\lambda$ can be estimated within its $\theta$ error with a probability of at least $1\!-\!\delta$.
That is, let $\lambda_i$ be a random variable that denotes the connectivity of a randomly generated net where $1\leq\lambda_i \leq p$ (since a net connects at least one part and at most $p$ parts).
Let $\lambda'=\frac{1}{|N|}\sum \lambda_i$ be the estimation for $\lambda$ where $|N|$ denotes the total number of nets obtained in the stochastic hypergraph. By Hoeffding inequality,
\begin{equation}
Pr[|\lambda' - \lambda| \geq \theta] \leq 2 \exp(\frac{-2|N|\theta^2}{(p-1)^2})
\end{equation}
\noindent 
To achieve $1-\delta$ confidence, 
\begin{equation}
2 \exp(\frac{-2|N|\theta^2}{(p-1)^2}) \leq \delta
\end{equation}
\noindent must be achieved. 
Hence, solving this equation for $|N|$ gives 
\begin{equation}
|N| \geq \frac{(p-1)^2}{2\theta^2}\ln\frac{2}{\delta}
\label{equ:hoefding-bound}
\end{equation}
\noindent which denote the smallest number of nets needed to achieve the $\theta$ error with $1\!-\!\delta$ confidence.

As shown, if an adequate number of nets are generated, the stochastic hypergraph model encodes the expected communication volume  with low error with high probability. Since the expected connectivity $\lambda$ is determined by the partitioning $\Pi$ over the hypergraph, the stochastic hypergraph partitioning can minimize this objective.
Additionally, if each vertex is equally likely to be selected in a mini-batch, then the same vertex weighting and balancing constraint in hypergraph model for full-batch training can be applied here to achieve computational load balance.

\subsection{Extension to GNNs}
The main difference between general GNN models~\cite{velivckovic2017graph, wu2019simplifying,klicpera2018predict} and GCNs is in the way messages are created and combined between vertices.
In some GNN models, DMM is performed first and messages are created, which is followed by a specialized SpMM for message passing and combining.
For example, in a GAT~\cite{velivckovic2017graph}, first each vertex feature is transformed with a local parameter matrix (i.e., DMM), and the resulting feature is transmitted to neighbor vertices using the same communication pattern as in SpMM. At the destination vertex, 
features are concatenated and then multiplied with an attention vector.  
That is, the order of SpMM and DMM can be changed and additional mathematical operations can be applied to their outputs, without affect the message directions and communication patterns between vertices.
Therefore, our proposed partitioning method can be directly used for other GNN models, and simple modifications to the proposed GCN algorithm can support the additional computations necessary alongside the same communication scheme as before.

\section{Experimental Results}
\label{sec:expt}

We evaluate the performance of the proposed parallel GCN training algorithm on a diverse set of real-world graphs from popular applications that use GCN models such as citation networks, social networks, road networks, and product co-purchasing networks.
Properties of these graphs are displayed in Table~\ref{table:dataset}.

\begin{table}[t]
\caption{Dataset properties}
\small	
\begin{tabular}{lrrl}
\toprule
Dataset   & \multicolumn{1}{c}{Vertices} & \multicolumn{1}{c}{Edges} & \multicolumn{1}{l}{Type} \\
\midrule
amazon0601 & 403,394 &	3,387,388 & Directed \\
cit-Patents & 3,774,768 & 16,518,948 &	Directed \\
coPapersDBLP & 540,486 & 30,491,458 &	Undirected \\
com-Amazon & 334,863 &	1,851,744 &	Undirected \\
com-Youtube & 1,134,890 &	5,975,248 &	Undirected \\
flickr & 820,878 &	9,837,214 &	Directed \\
roadNet-CA & 1,971,281 &	5,533,214 & Undirected \\
soc-Slashdot0902 & 82,168 &	948,464 &	Directed \\
\midrule
{Cora} & {2708} & {10556} & {Undirected} \\
{ogbn-Papers100M} & {111,059,956} & {1,615,685,872} & {Directed}  \\
{Reddit} & {232,965} & {114,615,892} & {Undirected} \\
\bottomrule
\end{tabular}
\label{table:dataset}
\end{table}


We use DGL~(with PyTorch v1.6 backend) implementation of GCN as the baseline, and compute speedup values according to its single-node CPU performance.
We also compare our performance against CAGNET~\cite{tripathy2020reducing} which is the algorithm most related to our own, by using both the original GPU implementation and our own CPU implementation of CAGNET. 
We omit comparisons against Neugraph~\cite{ma2019neugraph} and ROC~\cite{jia2020improving} as they are not compatible with CPU clusters, and CAGNET already provides much more scalability.
To the best of our knowledge, our algorithm is the first parallel GCN training algorithm designed for CPU clusters.


We evaluate the improvements in performance of the proposed parallel GCN training algorithm with both hypergraph partitioning (HP) and graph partitioning (GP) models used to partition the input matrices.
We also evaluate our novel stochastic hypergraph partitioning (SHP) model for mini-batching.
Additionally, we report results for random partitioning (RP) as a baseline, which evenly splits the adjacency matrix by assigning rows to processors uniformly at random, and is a competitive method for balancing computational load and communications.


We run our CPU experiments on a cluster of 180 compute nodes with 2x Intel Xeon Platinum 8268 2.9 GHz 24-core processors~(48~cores per node) and 4GB RAM per core. 
Our GPU experiments use the Sulis cluster of 30 nodes each with 3x NVIDIA A100 GPUs and 4GB RAM per core.
Both use InfiniBand interconnect (100 Gbit/s) and Slurm Workload Manager.
The single-node DGL implementation requires a server with a better hardware configuration. 
We use a 16-core Intel Xeon 3.90GHz processor with 500 GB memory.

Our CPU code is in C++, using SuiteSparse:GraphBLAS library for local sparse matrix operations and MPI for point-to-point communication operations.
The GPU version in Python uses PyTorch with NCCL backend to perform communication operations~\cite{awan2016efficient}.
Incorporating future support for asynchronous communication (as in our CPU implementation) may help overcome the limitations of NCCL to overlap communication and computation for better performance gains on GPU. 
We used PaToH~\cite{ccatalyurek2011patoh} hypergraph partitioning tool and METIS~\cite{karypis1998hmetis} graph partitioning tool. 
For \texttt{ogbn-Papers100M}, we used KaHyPar~\cite{kahypar} which can handle massive-scale graphs.
We used both partitioning tools with their default parameters and set the maximum imbalance ratio as $\epsilon\!=\!0.01$.

\begin{table}[t]
\renewcommand{\arraystretch}{0.9}
\caption{Performance comparison with HP, GP, and RP on $P\!=\!512$ processors} 
\label{tables:table1}
\centering
\setlength\tabcolsep{3pt}
\small	
\begin{tabular}{l*7r}
& & & \multicolumn{2}{c}{Volume} & \multicolumn{2}{c}{Messages} & \\
\cmidrule(r){4-5}\cmidrule(r){6-7}
& & \multicolumn{1}{c}{R}   & \multicolumn{1}{c}{Avg} & \multicolumn{1}{c}{Max}         & \multicolumn{1}{c}{Avg} & \multicolumn{1}{c}{Max} & \multicolumn{1}{c}{S} \\
\midrule
\multirow{3}{*}{amazon0601}       & HP & 0.63  & 0.12 & 0.29 & 0.22 & 0.51 & 10.88  \\
                                  & GP & 0.65  & 0.18 & 0.31 & 0.30 & 0.62 & 10.55  \\
                                  & HP/GP & 0.97  & 0.67 & 0.92 & 0.74 & 0.82 &       \\
\midrule
\multirow{3}{*}{cit-Patents}      & HP & 0.77  & 0.17 & 0.29 & 0.70 & 0.89 &  8.48  \\
                                  & GP & 0.80  & 0.19 & 0.50 & 0.77 & 0.94 & 8.10  \\
                                  & HP/GP & 0.95  & 0.88 & 0.57 & 0.91 & 0.95 &       \\
                
\midrule
\multirow{3}{*}{coPapersDBLP}     & HP & 0.32  & 0.07 & 0.08 & 0.42 & 0.71 & 10.93  \\
                                  & GP & 0.69  & 0.07 & 0.16 & 0.57 & 0.77 & 5.04  \\
                                  & HP/GP & 0.46  & 0.97 & 0.49 & 0.74 & 0.92 &       \\
\midrule
\multirow{3}{*}{com-Amazon}       & HP & 0.32  & 0.09 & 0.20 & 0.14 & 0.32 & 14.31 \\
                                  & GP & 0.37  & 0.14 & 0.27 & 0.19 & 0.42 & 12.37  \\
                                  & HP/GP & 0.86  & 0.60 & 0.73 & 0.72 & 0.75 &       \\
                                  
\midrule
\multirow{3}{*}{com-Youtube}      & HP & 0.40  & 0.36 & 0.52 & 0.72 & 0.97 & 10.85  \\
                                  & GP & 1.45  & 0.37 & 2.60 & 0.90 & 0.99 & 3.01  \\
                                  & HP/GP & 0.28  & 0.98 & 0.20 & 0.81 & 0.98 &       \\
                                  
\midrule
\multirow{3}{*}{flickr}           & HP & 0.81  & 0.45 & 0.60 & 0.79 & 1.00 & 9.59  \\
                                  & GP & 11.13 & 0.38 & 6.89 & 0.96 & 1.00 & 0.70  \\
                                  & HP/GP & 0.07  & 1.19 & 0.09 & 0.82 & 1.00 &       \\
                                  
\midrule
\multirow{3}{*}{roadNet-CA}       & HP & 0.19  & 0.01 & 0.01 & 0.01 & 0.03 & 30.32 \\
                                  & GP & 0.20  & 0.01 & 0.02 & 0.01 & 0.03 & 29.08 \\
                                  & HP/GP & 0.96  & 0.78 & 0.67 & 1.03 & 1.00 &       \\
                
\midrule
\multirow{3}{*}{soc-Slashdot0902} & HP & 0.75  & 0.74 & 0.69 & 0.86 & 0.92 & 3.50  \\
                                  & GP & 2.02  & 0.85 & 4.38 & 0.93 & 1.00 & 1.30  \\
                                  & HP/GP & 0.37  & 0.86 & 0.16 & 0.92 & 0.92 &       \\

\midrule
\multicolumn{2}{r}{mean HP}             & 0.47  & 0.13 & 0.21 & 0.29 & 0.48 & 10.60  \\
\multicolumn{2}{r}{mean GP}             & 0.98  & 0.15 & 0.56 & 0.35 & 0.52 & 5.04  \\
\multicolumn{2}{r}{mean HP/mean GP}     & 0.48  & 0.87 & 0.37 & 0.83 & 0.92 &   \\ 
\midrule
\end{tabular}
\end{table}

\begin{figure*}[t]
\centering
\includegraphics[width=0.3\textwidth]{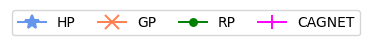}
\includegraphics[width=\textwidth]{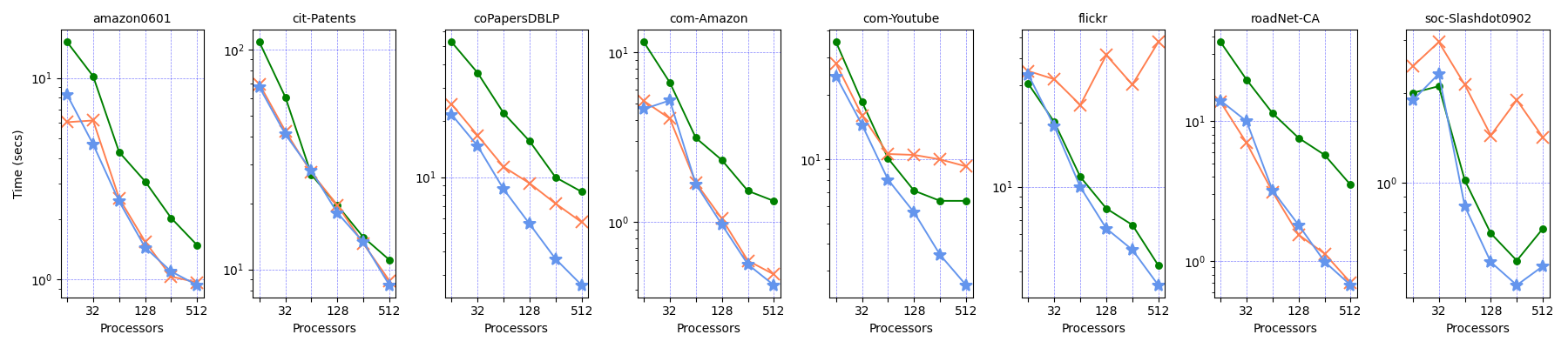}
\includegraphics[width=\textwidth]{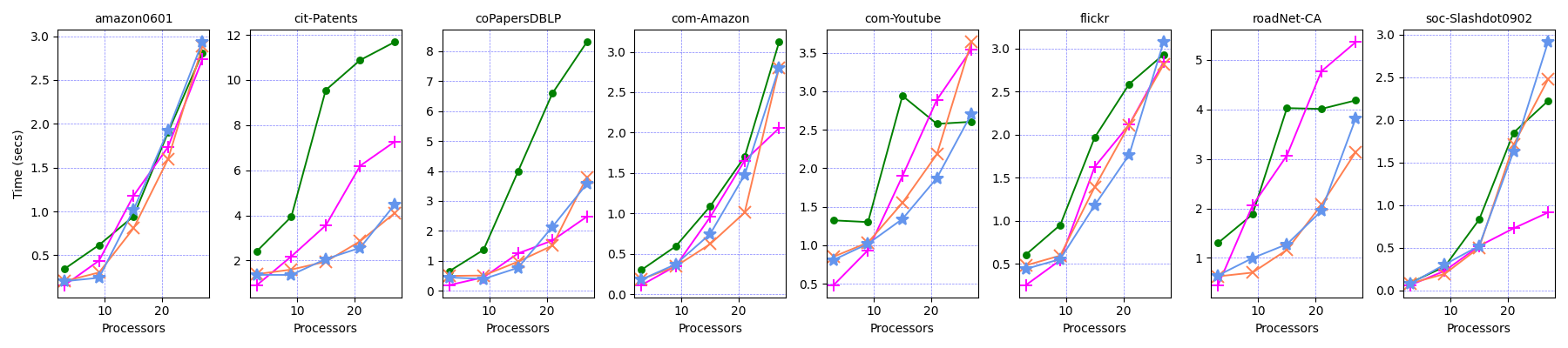}
\caption{Strong scaling for full-batch training with HP, GP, and RP on $P\!=\!16$ to $P\!=\!512$ CPUs (top row) and with HP, GP, RP, and CAGNET (CN) on $P=3$ to $P=27$ GPUs (bottom row).}
\label{fig:strong-scaling}
\end{figure*}

\textbf{Communication Costs.} Table~\ref{tables:table1} compares HP, GP, and RP in terms of the communication volume and message counts metrics they incur on $512$ processors~(i.e., MPI processes).
For each partitioning method, we ran the parallel GCN algorithm with random vertex features and label data for five epochs and measured the running time average and total communication cost metrics.
These metrics respectively relate to bandwidth and latency costs induced by different partitioning strategies on parallelization costs.
In the table, both the average and maximum volume/number of messages sent by a processor are displayed.
Average values are proportional to the total message volume/count values, and are used to display how much the maximum values deviate from the mean.

For each input graph, the first and second rows denote the respective values attained by HP and GP, where these values are normalized with respect to values attained by RP.
The third row denotes the ratios of values attained by HP and GP~(i.e., HP/GP).
The first column~(i.e., ``R'') in the table indicates the ratio of the parallel running time of HP and GP to that of RP.
The last column~(i.e., ``S'') denotes the speedup values attained by HP and GP with respect to single-node running time performance of DGL.
At the end of the table, the geometric means of the normalized values for HP and GP are given where the ratios of these values are given in the last row.
For instance, the ``R'' and ``S'' columns for \texttt{amazon0601} are interpreted as follows: Parallel running times of HP and GP divided by that of RP is $0.63$ and $0.65$, respectively.
The parallel running time of HP divided by that of GP is $0.97$.
Speedups achieved by HP and GP with respect to DGL are displayed under column ``S'' as $10.88$ and $10.55$, respectively.

As seen in Table~\ref{tables:table1}, both HP and GP provide significant improvements over communication volume and message count metrics.
On average, HP and GP incur $87\%$ and $85\%$ less average communication volume than RP respectively, with HP performing $15\%$ better than GP.
In terms of maximum communication volume, HP consistently outperforms RP, providing $79\%$ improvement on average over RP.
HP performs $63\%$ better than GP and provides better communication balance.
Even though GP provides $44\%$ improvement on average over RP, its performance significantly degrades for graphs \texttt{com-Youtube}, \texttt{flickr} and \texttt{soc-Slashdot0902} where for instance, GP performs $6.89$x worse than RP for \texttt{flickr}.
Although both partitioning methods provide significant improvement in average communication volume, graph partitioning can disrupt the communication balance between processors.
For the message count metrics, on average, HP and GP reduce the total number of messages by $71\%$ and $65\%$ as compared to RP while HP performs $17\%$ better than GP.
Similarly, maximum message count is respectively reduced by $52\%$ and $48\%$ by HP and GP while HP performs $9\%$ better than GP.

\begin{figure*}[t]
    \centering
    \begin{subfigure}[b]{0.43\textwidth}
    \centering
    \includegraphics[width=\textwidth]{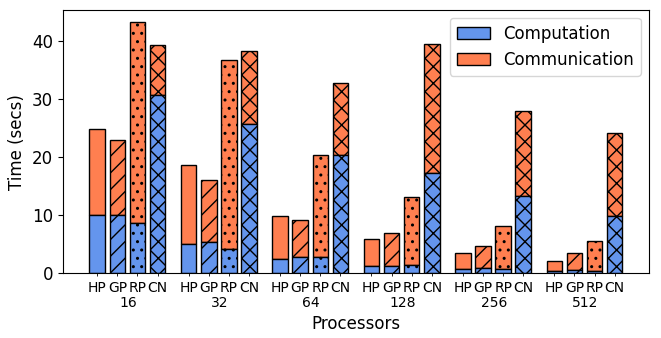}
    \phantomsubcaption
    \label{fig:barchart}
    \end{subfigure}
    \begin{subfigure}[b]{0.38\textwidth}
    \centering
    \includegraphics[width=\textwidth]{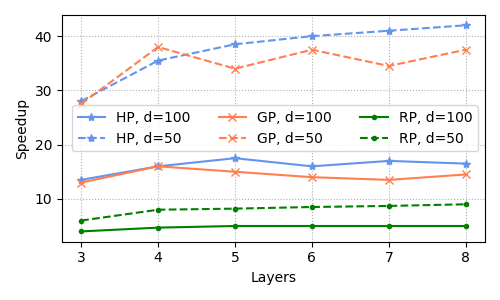}
    \phantomsubcaption
    \label{fig:perf-var}
    \end{subfigure}
    \begin{subfigure}[b]{0.18\textwidth}
    \centering
    \includegraphics[width=\textwidth]{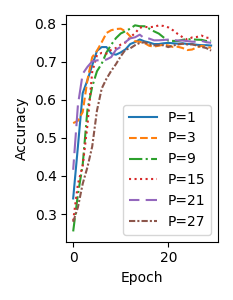}
    \phantomsubcaption
    \label{fig:accuracy}
    \end{subfigure}
    \caption{Performance comparisons for full-batch training. \subref{fig:barchart} Communication time and computation time split with HP, GP, RP, and CAGNET (CN) on \texttt{coPapersDBLP} for $P\!=\!16$ to $P\!=\!512$ CPUs.
    \subref{fig:perf-var} Speedup with increasing layers ($L=3,4,\ldots,8$) and dimensions ($d=50,100$) on \texttt{roadnet-CA} for $P=512$ CPUs.
    \subref{fig:accuracy} GNN model accuracy with HP on \texttt{Cora} for $P=1$ to $P=27$ GPUs.}
\end{figure*}

\textbf{Parallel Running Times.}
Improvements in communication costs by HP and GP considerably reduce parallel running of RP.
On all graphs, HP provides an average of $2.12$x speedup over RP.
GP runs $20\%$--$80\%$ faster than RP for most graphs. However, it is slower than RP on \texttt{flickr}, \texttt{com-Youtube} and \texttt{soc-Slashdot0902}.
The reason for this is the communication volume imbalance, as can be seen from the maximum and average communication volumes achieved on these graphs.
Note that RP achieves a good communication and computation balance.
The best performance is achieved by HP and GP for \texttt{roadNet-CA} where both partitioning methods provide approximately $99\%$ improvement in communication volume and message count metrics compared to RP and runs $5$x faster.

As seen in the speedup column, on average, HP and GP provide $10.60$x and $5.04$x speedup respectively over the DGL implementation. 
The best speedup is achieved for \texttt{roadNet-CA} where HP and GP approximately provide $30.32$x and $29.08$x speedup.
This is because road networks are relatively more sparse as compared to the other social networks and hence the amount of data transferred between processors reduces in such cases.
As the graph sizes increase and graphs become more sparse, partitioning tools usually perform better optimizations.

Figure~\ref{fig:strong-scaling} displays strong scaling of \mbox{HP}, \mbox{GP}, and \mbox{RP} on CPU~(first row) and GPU~(second row) clusters.
As seen here, on the CPU cluster, HP achieves almost linear speedup up to $512$ cores on all input graphs.
Additionally, HP either matches or outperforms GP, and always outperforms RP.
The reason for the speedup loss of HP on 512 processors for \texttt{soc-Slashdot0902} is the relatively smaller size of the graph as compared to the others.
In general, GP performs better than RP except for \texttt{flickr}, \texttt{com-Youtube} and \texttt{soc-Slashdot0902} graphs due to degradation of communication balance as also shown in Table~\ref{tables:table1}.
Here, we omit plot for our CPU implementation of CAGNET since HP and GP significantly outperform it.

We also demonstrate that algorithms that are focused on optimizing communication operations are better suited to CPU clusters over GPUs, and where the sparsity of the problem is important to exploit.
As seen in Figure~\ref{fig:strong-scaling},
for GPU cluster experiments, the PyTorch implementations~(with NCCL backend) of both CAGNET and our parallel GCN~(i.e. HP, GP and RP) do not scale well (up to $27$ GPUs). 
The reason for this is the high communication efficiency needed to attain speedup on GPUs. In comparison to MPI, the NCCL backend cannot provide the necessary efficiency.
On GPUs, the proportion of total running time that is spent on local computation is small, therefore the gains obtained via parallelization do not amortize the time spent for communication on larger GPU counts.
In addition, despite the optimizations obtained in the communication volume from our algorithm, with the NCCL backend these are not as effective as with MPI. This results in limited performance improvement on the overall parallel run time due to the higher latency costs.
HP and GP continue to be faster than CAGNET for most datasets and settings.
In addition, the performance improvement is expected to be more stark at higher GPU counts, as can be seen from the CPU cluster results, but no suitable larger GPU clusters were available for our experiments.
Moreover, we find that our parallel CPU implementation for HP is able to outperform the GPU version in many cases. For example, on \texttt{amazon0601}, the running time is $0.94$ seconds on $512$ CPUs, while on $15$ GPUs it takes as long as $1.02$ seconds. On the larger \texttt{roadNet-CA} dataset, the same setting takes only $0.67$ seconds on CPU and twice as long ($1.28$ seconds) on GPU.

\textbf{Communication and Computation Times.}
Figure~\ref{fig:barchart} analyses the breakdown of communication and local computation times in the total parallel CPU running time of HP, GP, RP, and CAGNET (CN) for \texttt{coPapersDBLP}.
On all processor counts, HP and GP consistently perform better than CAGNET, with HP being the best method at high processor counts. 
On the largest processor count, HP 
runs nearly $12$x faster than CAGNET.
Even though RP performs worse than CAGNET on $P\!=\!16$ processors, its performance becomes better as the number of processors increases.
As seen in the figure, the total communication time decreases with the total computation time for HP, GP and RP as the number of processors increases, whereas the communication time of CAGNET increases.
This is because point-to-point communication necessitates each processor to communicate only with a small subset of processors and thus incurs lower communication volume and latency costs, whereas broadcast communication involves all processors and incurs higher communication overheads due to the unnecessary data and message transfer.
The redundant computations in CAGNET are also visible in its higher local computation times.
Moreover, better optimizations are achieved by HP than GP, which is evident from the communication time of GP being $1.7$x higher and CAGNET being $8.3$x higher than that of HP on $512$ processors. HP shows between $2.4$x to $3$x better communication efficiency over RP from low processor counts to high, while the communication benefit of GP over RP drops slightly from $2.7$x to $1.8$x.

\textbf{Scalability for Deeper Networks.}
Figure~\ref{fig:perf-var} shows speedup performance of HP, GP, and RP when varying the number of layers and the dimensionality of features.
The number of dimensions is chosen as $d=50$ and $d=100$, and the number of layers is increased from $2$ to $8$.
The speedup 
is computed by dividing the running time of DGL by the running time of HP, GP, or RP respectively under the same GCN configuration.
When the number of layers increases and $d$ is kept constant, there is no loss of speedup in any algorithm, and speedup in fact increases for HP.
The speedups decrease as $d$ increases because of the rise in total communication volume, which reduces the parallelization efficiency.
For example, the speedup of HP decreases approximately from $40$x to $17$x when the number of features is increased from $d=50$ to $d=100$, for an $8$-layer GCN.
On the other hand, the performance of HP increases from approximately $28$x to $40$x when the number of layers is increased from $3$ to $8$, for $d=100$.
We observe the same behavior across different datasets, on both CPU and GPU versions of our algorithm, and present the plots on \texttt{roadNet-CA} for $512$ CPUs.

\textbf{Predictive Performance.} We also examine the effect on predictive performance of the GCN model when parallelized using our training algorithm. We use the \texttt{Cora} dataset since our large-scale networks do not have training labels.
We run the parallel training algorithm for 30 epochs on up to 27 GPUs, and compare their accuracy performance with the serial training algorithm.
Figure~\ref{fig:accuracy} shows that the parallel training algorithm does not have any negative impact on the accuracy performance, with approximately 75\% accuracy achieved in all settings.

\begin{figure}[tb]
\centering
\includegraphics[width=0.8\linewidth]{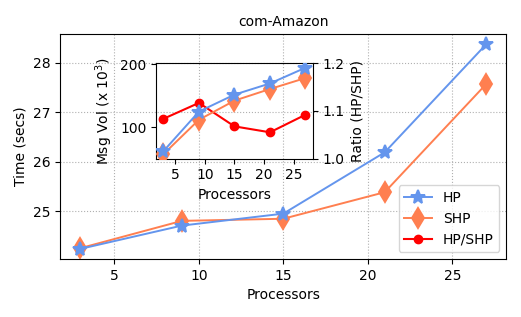}
\caption{Performance comparisons for mini-batch training. Running time and communication volume (``Msg Vol'') with HP and SHP on \texttt{com-Amazon} for $P=3$ to $P=27$ GPUs.}
\label{fig:stoch}
\end{figure}  

\textbf{Stochastic Hypergraph Model.}
Figure~\ref{fig:stoch} shows the relative performance improvement of stochastic hypergraph model~(SHP) over HP in mini-batch training.
10K random mini-batches of size 20K vertices are generated. The total communication volume they induce under partitionings obtained by HP and SHP on \texttt{com-Amazon} graph using GPU are measured.
We set $\theta\!=\!0.1$ and $\delta\!=\!0.5$ to run SHP and we set the same maximum imbalance ratio ($\epsilon=0.01$) for both SHP and HP.
In the figure (inset), the relative improvement of SHP over HP (in red along the secondary y-axis) shows that HP induces 10\% more communication volume than SHP on average. 
The performance difference in favor of SHP is even more pronounced at higher processor counts.
We see that SHP provides a greater benefit of shorter running time with more processors.

\textbf{Scalability to Billion-scale Datasets.}
We also test our algorithm on a billion-scale \texttt{ogbn-Papers100M} dataset, which is only feasible when partitioned onto $27$ GPUs due to memory limitations. Table~\ref{tab:100M} shows that our methods is scalable not only to high processor counts but also to very large graphs. RP slows down significantly with increasing dimensionality of features. On the other hand, the communication benefit of HP, reducing communication volume approximately by a factor of $10$x, allows it to scale better.

\begin{table}[tb]
\caption{Performance comparison with HP and RP ($d=1, 2, 5$) on \texttt{ogbn-Papers100M} for $P=27$ GPUs.}
\small	
\begin{tabular}{c|rrr|c}
\toprule
\multirow{2}{*}{\parbox{2cm}{\centering {Partitioning model}}}     & \multicolumn{3}{c|}{{Running time (secs)}} & \multirow{2}{*}{\parbox{2cm}{\centering {Communication volume}}} \\
                & {$d=1$} & {$d=2$} & {$d=5$} & \\
\midrule
{HP}              & {24.46} & {25.00} & {29.73} & {1.2 billion} \\
{RP}              & {34.70} & {42.88} & {65.14} & {13 billion} \\
\bottomrule
\end{tabular}
\label{tab:100M}
\end{table}

\textbf{Comparison against SOTA.}
Our optimizations are proposed for large-scale CPU clusters, since the improvement of point-to-point communication overheads over broadcast is more pronounced in such cases.
We nonetheless compare the running time of our GPU implementation~(HP) against state-of-the-art distributed GPU systems. All systems use the same GCN architecture and report results on the \texttt{Reddit} dataset that is common among them.
The reported algorithms (except CAGNET) use methods that affect training and predictive performance, such as caching, vertex replication, and asynchronous parameter updates, whereas HP performs full-batch training.
As seen from the results, HP achieves considerable relative performance even on small GPU counts. 


\begin{table}[tb]
\caption{Comparison of running time (per epoch) on \texttt{Reddit}.}
\small	
\begin{tabular}{lrll}
\toprule
{Method}   & \multicolumn{1}{c}{{Running time (per epoch)}} & \multicolumn{1}{c}{{Setup}} & \multicolumn{1}{l}{{Reference}} \\
\midrule
{HP} & {0.67} & {A100*3} & {-} \\
{CAGNET} & {0.11} & {V100*4} & {Fig~1 (c=1)~\cite{tripathy2020reducing}}\\
{ROC}  & {1/5 = 0.20} & {P100*4} & {Fig~5~\cite{jia2020improving}} \\
{Sancus}  & {97.4/1000 = 0.09} & {V100*4} & {Table~4~(SCS-A)~\cite{peng2022sancus}} \\
{PaGraph}  & {$\approx$ 1.00} & {1080Ti*1} & {Fig 9~\cite{lin2020pagraph}} \\
{Dorylus}  & {162.9/120 = 1.36} & {V100*2} & {Fig~5, Table~4~\cite{thorpe2021dorylus}} \\
{DGCL}  & {0.15} & {V100*4} & {Fig~8(a)~\cite{cai2021dgcl}} \\
\bottomrule
\end{tabular}
\label{table:sota}
\end{table}



\section{Conclusion}

We proposed a highly parallel algorithm for GCN training on large-scale distributed-memory systems.
For scalability, 
all matrices except parameter matrices are row-wise partitioned between processors.
The algorithm achieves further communication cost reduction by capturing the sparsity pattern of the adjacency matrix to perform point-to-point communications, via the use of a sparse matrix partitioning scheme based on an intelligent hypergraph model.

Our solution is scalable on a CPU cluster with MPI backend, with the proposed hypergraph partitioning based approach providing significant speedups.
The latency between hidden layers are considerably amortized, allowing deeper GCN models to be trained, and with no impact on accuracy.
We also performed experiments on a GPU cluster with NCCL backend which provide useful insights on large scale GNN training. All tested algorithms demonstrated less scalability in GPUs compared to the CPU based versions.
We also observed that on some instances CPU implementation runs faster than the GPU implementation, besides being more scalable.
To further improve the mini-batch training, we proposed a novel stochastic hypergraph model that successfully captures the randomness of communication operations in parallel mini-batch training and achieves improvements over the hypergraph model.

The proposed algorithm is adaptable to other GNNs by changing only the local computations without requiring any changes in terms of the communication operations, which opens up many future directions of research.
Additionally, there is scope for exploring several optimizations in the mini-batch sampling strategy and in the GPU communications.
The use of 2D and 3D partitioning schemes is another promising avenue for further research.






\begin{acks}
Aparajita is supported by the Feuer International Scholarship in Artificial Intelligence. Computing resources used were provided by the Scientific Computing Research Technology Platform at the University of Warwick.
\end{acks}


\balance
\bibliographystyle{ACM-Reference-Format}
\bibliography{paper}

\end{document}